%% file: ms.tex
\documentclass[letterpaper]{article} % DO NOT CHANGE THIS
\usepackage{aaai20}  % DO NOT CHANGE THIS
\usepackage{times}  % DO NOT CHANGE THIS
\usepackage{helvet} % DO NOT CHANGE THIS
\usepackage{courier}  % DO NOT CHANGE THIS
\usepackage[hyphens]{url}  % DO NOT CHANGE THIS
\usepackage{graphicx} % DO NOT CHANGE THIS
\usepackage{graphicx}  % DO NOT CHANGE THIS
\usepackage[utf8]{inputenc} % allow utf-8 input

\usepackage{placeins}

\usepackage{booktabs}       % professional-quality tables
\usepackage{amsfonts}       % blackboard math symbols
\usepackage{graphicx}       % to load images
\usepackage[subpreambles=true]{standalone}

\usepackage{amsfonts,amsmath,amssymb,amsthm,mathtools} % various math stuff

\newcommand{\trp}{{^\top}} % transpose

\renewcommand{\eqref}[1]{eq.~\ref{eq:#1}}
\newcommand{\Nrm}{\mathcal{N}}

\newcommand{\figref}[1]{Fig.~\ref{fig:#1}}  % use for citing figs
\newcommand{\secref}[1]{Sec.~\ref{sec:#1}}  % use for citing secs
\newcommand{\algoref}[1]{Algorithm~\ref{algo:#1}}  % use for citing algorithms

\newcommand{\suppsecref}[1]{Appendix ~\ref{supp:#1}}  

\newcommand{\vx}{\mathbf{x}}

\newcommand{\Dat}{\mathcal{D}}

\newcommand{\vb}{\mathbf{b}}

\newcommand{\vphi}{\mathbf{\ensuremath{\bm{\phi}}}}

\newcommand{\vm}{\mathbf{m}}

\newcommand{\vw}{\mathbf{w}}
 
\newcommand{\vs}{\mathbf{s}}

\newcommand{\vtheta}{\mathbf{\ensuremath{\bm{\theta}}}}
\newcommand{\LL}{\ensuremath{\mathcal{L}}}

\newcommand{\mR}{\mathbf{R}}
\newcommand{\mW}{\mathbf{W}}

\newcommand{\vh}{\mathbf{h}}

\newtheorem{thm}{Theorem}[section]

\newtheorem{prop}{Proposition}

\def\Pr{\ensuremath{\text{Pr}}}

\usepackage{mdwlist}  % Make list items closer together: itemize*, enumerate*
\usepackage{bm,amsbsy} % for bold symbols
\usepackage{algorithm,algorithmic}

\usepackage[capitalise]{cleveref}

\usepackage{pgfmath}
\usepackage{pgf}
\usepackage{tikz}
\usepackage{pgfplots}
	\usepgfplotslibrary{groupplots}
    \usetikzlibrary{calc}
    \usetikzlibrary{positioning}
    \usetikzlibrary{angles,quotes}
    \usetikzlibrary{backgrounds}
    \usetikzlibrary{external}
    \usetikzlibrary{fit}
    \usetikzlibrary{bayesnet}
    \usetikzlibrary{calc}
    \usetikzlibrary{fit}
    \usetikzlibrary{spy, shadows}
    \usetikzlibrary{arrows.meta}
    \usetikzlibrary{shadings}
    \usetikzlibrary{fadings}
    \usetikzlibrary{matrix}
    
\newcommand*\circled[2]{\tikz[baseline=(char.base)]{
            \node[shape=circle,draw, #1] (char) {#2};}}
\newcommand{\citet}[1]{\citeauthor{#1} \shortcite{#1}}
\newcommand{\citep}{\cite}

\usepackage{xcolor}
	\definecolor{lightgray}{gray}{0.95}
    \definecolor{C1}{HTML}{1F77B4}
    \definecolor{C2}{HTML}{FF7F0E}
    \definecolor{C3}{HTML}{2CA02C}
    \definecolor{C4}{HTML}{D62728}
    \definecolor{C5}{HTML}{9467BD}
    \colorlet{C1light}{C1!20!white}
    \colorlet{C2light}{C2!20!white}
    \colorlet{C3light}{C3!20!white}
    \colorlet{C4light}{C4!20!white}
    \colorlet{C5light}{C5!20!white}
    \definecolor{darkblue}{rgb}{0.0, 0.0, 0.55}
    \definecolor{darkred}{rgb}{0.55, 0.0, 0.0}

\usepackage[colorinlistoftodos,prependcaption,textsize=tiny]{todonotes}
\usepackage{xargs}
\newcommandx{\mpnote}[2][1=]{\todo[linecolor=red,backgroundcolor=red!25,bordercolor=red,#1]{MP: #2}}
\newcommandx{\fnote}[2][1=]{\todo[linecolor=blue,backgroundcolor=blue!25,bordercolor=blue,#1]{F: #2}}
\newcommandx{\mbnote}[2][1=]{\todo[linecolor=green,backgroundcolor=green!25,bordercolor=green,#1]{MB: #2}}
\newcommandx{\thiswillnotshow}[2][1=]{\todo[disable,#1]{#2}}

\newcommand\blfootnote[1]{%
  \begingroup
  \renewcommand\thefootnote{}\footnote{#1}%
  \addtocounter{footnote}{-1}%
  \endgroup
}

\title{Interpretable and Differentially Private Predictions}
\author{Frederik Harder,\textsuperscript{\rm 1}\textsuperscript{\rm 2} Matthias Bauer,\textsuperscript{\rm 1}\textsuperscript{\rm 3} Mijung Park\textsuperscript{\rm 1}\textsuperscript{\rm 2}\\ 
\textsuperscript{\rm 1} Max Planck Institute for Intelligent Systems, Tübingen, Germany\\
\textsuperscript{\rm 2} University of Tübingen, Tübingen, Germany\\
\textsuperscript{\rm 3} Department of Engineering, University of Cambridge, Cambridge, UK\\
\{fharder$|$bauer$|$mpark\}@tue.mpg.de
}

\begin{document}

\maketitle

\begin{abstract}
Interpretable predictions, where it is clear why a machine learning model has made a particular decision, can compromise privacy by revealing the characteristics of individual data points. 
This raises the central question addressed in this paper: \emph{Can models be interpretable without compromising privacy?} 
For complex ``big'' data fit by correspondingly rich models, balancing privacy and explainability is particularly challenging, such that this question has remained largely unexplored. 
In this paper, we propose a family of simple models in the aim of approximating complex models using several \emph{locally linear maps} per class to provide high classification accuracy, as well as differentially private explanations on the classification. 
We illustrate the usefulness of our approach on several image benchmark datasets as well as a medical dataset.
\end{abstract}

\section{Introduction}

\blfootnote{This work was published at AAAI 2020. Please check \url{https://aaai.org/Library/AAAI/aaai-library.php} for the published version.}

\begin{figure}%{r}{0.5\textwidth}
    \centering
    % \vskip-1em
    \begin{tikzpicture}
        \node[rounded corners, fill=lightgray] (int) {Interpretability};
        \node[rounded corners, fill=lightgray, below left=.8cm and 0.5cm of int] (privacy) {Privacy};
        \node[rounded corners, fill=lightgray, below right= .8cm and 0.5cm of int] (accuracy) {Accuracy};
        \draw[thick, draw=darkgray, stealth'-stealth'] (int.east) -- (accuracy.north) node[pos=0.5, above right] {\circled{inner sep=2pt}{B}};
        \draw[thick, draw=darkgray, stealth'-stealth'] (privacy.east) -- (accuracy.west)  node[pos=0.5, above] {\circled{inner sep=2pt}{C}};
        \draw[thick, draw=darkgray, stealth'-stealth'] (int.west) -- (privacy.north) node[pos=0.5, above left] {\circled{inner sep=2pt}{A}};
    \end{tikzpicture}
    \caption{Modern machine learning systems need to trade off accuracy, privacy, and interpretability.}
    % \vskip-1em
    \label{fig:LoveTriangle}  % :D
\end{figure}
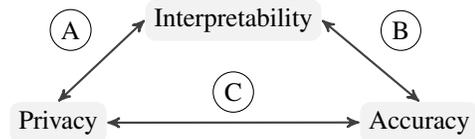

The \textit{General Data Protection Regulation (GDPR)} by the European Union imposes two important requirements on algorithmic design, \textit{interpretability} and \textit{privacy} \cite{Voigt2017}. These %As the GDPR's 
requirements introduce new standards on future algorithmic techniques, making them %these issues are
of particular concern to the machine learning community \cite{2016arXiv160608813G}. This paper addresses these two requirements in the context of classification, and studies the trade-off between privacy, accuracy and interpretability, see \figref{LoveTriangle}. 

Broadly speaking, there are two options to take for gaining interpretability: (i) rely on \emph{inherently interpretable models}; and (ii) rely on \emph{post-processing schemes} to probe trained complex models. 
Inherently interpretable models are often relatively simple and their predictions can be easily analyzed in terms of their respective input features. %Models that fall into this category are, f
For instance, in logistic regression classifiers and sparse linear models the coefficients represent the importance of each input feature. 
However, modern ``big'' data typically exhibit complex patterns, such that these relatively simplistic models often have lower accuracy than more complex ones. 

In order to soften this trade-off between interpretability and accuracy (\figref{LoveTriangle} \circled{inner sep=1pt}{B})
many post-processing schemes aim to gain insights from complex models like deep neural networks. One prominent aspect of this approach are gradient-based attribution methods \cite{SelvarajuDVCPB16,Ribeiro2016,SmilkovTKVW17,SundararajanTY17,MontavonBBSM15,journal.pone.0130140,abs-1711-06104}.

% \sout{When it comes to privacy,} 
In another line of research, many recent papers address the concern that complex models with outstanding predictive performance can expose sensitive information from the dataset they were trained on \cite{Carlini_et_al_2018,Song17,Shokri15,Fredrikson15}.
To quantify privacy, many recent approaches adopt the notion of 
%Many existing methods adopt the gold-standard notion of privacy, called 
{\it{differential privacy}} (DP), which provides a mathematically provable definition of privacy, and can quantify the level of privacy an algorithm or a model provides 
\cite{Dwork14}. 
In plain English, an algorithm is called differentially private (DP), if its output is random enough to obscure the participation of any single individual in the data. 
The randomness is typically achieved by injecting %typically comes from an injection of 
noise into the algorithm. The amount of noise is determined by the level of privacy the algorithm guarantees and the sensitivity, a maximum difference in its output depending on a single individual's participation or non-participation in the data (See \secref{background} for a mathematical definition of DP). 

There is, however, a natural trade-off between privacy and accuracy (\figref{LoveTriangle} \circled{inner sep=1pt}{C}): 
A large amount of added noise provides a high level of privacy but also harms prediction accuracy. When the number of parameters is high, like in deep neural network models, juggling this trade-off is very challenging, as privatizing high dimensional parameters results in a high privacy loss to meet a good prediction level 
\cite{2016arXiv160700133A}. 
Therefore, most existing work in differential privacy literature considers relatively small networks or assumes that some relevant data are publicly available to train a significant part of the network without privacy violation to deal with the trade-off.  %to keep the privacy loss for training the entire model moderately low 
(See \secref{relatedwork} for details). However, none of the existing work takes into account the interpretability of the learned models and this is our core contribution described below.

%%%%%%%%%%%%%%%%%%%%%%%%%%%%%%%%%%%%%%%%%%%%
\paragraph{Our contribution} 
In this paper, we study the trade-off between interpretability, privacy, and accuracy by making the following three contributions. %, i.e., we address all three arrows in \figref{LoveTriangle}.
\begin{itemize}
    \item \textbf{Proposing a novel family of interpretable models}:
To take into account privacy \emph{and} interpretability (\figref{LoveTriangle} \circled{inner sep=1pt}{A}), we propose a family of inherently interpretable models that 
can be trained privately. These models approximate the mapping of a complex model from the input data to class score functions, using several \textit{locally linear maps} (LLM) per class. 
Our formulation for LLM is inspired by the approximation of differentiable functions as a collection of piece-wise linear functions, i.e., the first-order Taylor expansions of the function at a sufficiently large number of input locations. 
Indeed, our local models with an adequate number of linear maps permit a relatively slight loss in accuracy compared to complex model counterparts\footnote{The level of loss in prediction accuracy depends on the complexity of data.}. 
    \item \textbf{Providing DP ``local" and ``global" explanations on classification}: Our model LLM, trained with the DP constraint,  provides insights on the key features %TODO
for classification at a ``local" and ``global" level. A \emph{local} explanation of a model illustrates how the model behaves at and around a specific input, showing how relevant different features of the input were to the model decision. 
This is a typical outcome one could obtain by probing a complex model using existing attribution methods. 
However, our model also provides a \emph{global} explanation, illustrating how the model functions as a whole and, in the case of classification, what types of input the different classes are sensitive to. This is what distinguishes our work from other existing post-processing attribution methods. 
    \item \textbf{Proposing to use random projections to better deal with privacy and accuracy trade-off}:
    We propose to adopt 
the \textit{Johnson-Lindenstrauss transform}, a.k.a., \textit{random projection} \cite{privacy-via-the-johnson-lindenstrauss-transform}, to decrease the dimensionality of each LLM 
and then privatize the resulting lower dimensional quantities. We found that exploiting data-indepdent random projection achieves a significantly better trade-off for high-dimensional image data. 
\end{itemize}

We would like to emphasize that our work is the first to address the interplay between interpretability, privacy, and accuracy. Hence, this work presents not only a novel inherently interpretable model but also an important conceptual contribution to the field that will spur more research on this intersection. 

%%%%%%%%%%%%%%%%%%%%%%%%%%%%%%%%%%%%%%%%%%%%

\section{Background on Differential Privacy}\label{sec:background}
% \mpnote{Taken from my other paper. Need to edit it to avoid self plagiarism}

We start by introducing differential privacy and a composition method that we will use in our algorithm, as well as  random projections. 

\paragraph{Differential privacy.}
Consider an algorithm $\mathcal{M}$ and neighboring datasets $\Dat$ and $\Dat'$ differing by a single entry, where the dataset $\Dat'$ is obtained by excluding one datapoint from the dataset $\Dat$. 
In DP \cite{Dwork14},
the quantity of interest is 
\emph{privacy loss}, defined by 
% of an outcome $o$ is defined by
% \vspace{-0.1cm}
\begin{equation}\label{eq:privacyloss}
L^{(o)} = \log \frac{\Pr(\mathcal{M}_{(\Dat)} = o)}{\Pr(\mathcal{M}_{(\Dat')} = o)},
% \vspace{-0.1cm}
\end{equation} where $\mathcal{M}_{(\Dat)}$ and $\mathcal{M}_{(\Dat')}$ denote the outputs of the algorithm given $\Dat$ and $\Dat'$, respectively. $\Pr(\mathcal{M}_{(\Dat)}=o)$ denotes the probability that $\mathcal{M}$ returns a specific output $o$. %, tells us how likely the event of the algorithm outputting a specific output $o$ can occur. 
When the two probabilities in \cref{eq:privacyloss} %(the numerator and denominator in \eqref{privacyloss})
are similar, even a strong adversary, who knows all the datapoints in $\Dat$ except for one, could not discern the one datapoint by which $\Dat$ and $\Dat'$ differ, based on the output of the algorithm alone. On the other hand, when the probabilities are very different, it  would be easy to identify the exclusion of the single datapoint in $\Dat'$. 
Hence, the privacy loss quantifies how revealing an algorithm's output is about a single entry's presence in the dataset $\Dat$. 
Formally, an algorithm $\mathcal{M}$ is called $\epsilon$-DP if and only if
$|L^{(o)}| \leq \epsilon, \forall o$.
A weaker version of the above is ($\epsilon, \delta$)-DP, if and only if
$|L^{(o)}| \leq \epsilon$, with probability at least $1-\delta$.  %TODO
%

%A popular way of designing DP algorithms is by introducing a noise addition step to the algorithm. 
Introducing a noise addition step is a popular way of making an algorithm DP.
The \textit{output perturbation} method achieves this by adding noise to the output $h$,
where the noise is calibrated to $h$'s {\it{sensitivity}}, defined by 
% denoted by $S_h$.
% A common form of sensitivity is the L2-sensitivity, which is the maximum difference in terms of L2-norm, under the one datapoint's difference in $\Dat$ and $\Dat'$, 
\begin{equation}\label{eq:sensitivity}
    S_h = \max_{\Dat, \Dat', |\Dat-\Dat'|=1} \|h(\Dat) - h(\Dat) \|_2,
\end{equation}  which is the maximum difference in terms of L2-norm, under the one datapoint's difference in $\Dat$ and $\Dat'$.
With the sensitivity, we can privatize 
the output using the \textit{Gaussian mechanism}, which simply adds Gaussian noise of the form: $\tilde{h}(\Dat) = h(\Dat) + \Nrm(0, S_h^2\sigma^2 \mathbf{I}_p),$
where $\Nrm(0,S_h^2\sigma^2 \mathbf{I}_p)$ means the Gaussian distribution with mean $0$ and covariance $S_h^2\sigma^2 \mathbf{I}_p$.
The resulting quantity $\tilde{h}(\Dat) $ is $(\epsilon, \delta)$-DP, where $\sigma \geq \sqrt{2\log(1.25/\delta)}/\epsilon$ (see \citet{Dwork14} for a proof). 
In this paper, we use the Gaussian mechanism to achieve differentially private LLM.

\paragraph{Properties of differential privacy.}
DP has two important properties: (i) post-processing invariance and (ii) composability. 
\textit{Post-processing invariance} states that applying any \textit{data-independent} mechanism to a DP quantity does not alter the privacy level of the resulting quantity. 
A formal definition of this property is given in \suppsecref{DPproperties}.  

\textit{Composability} states that combining DP quantities degrades privacy. 
% For instance, if one computes a statistic given a dataset and adds Gaussian noise, and repeats this routine multiple times, combining these privatized quantities, the average of these quantities will be quite close to the true statistic. 
% %
% Hence, one needs to increase the noise level to keep the strength of the privacy guarantee after the repeated use of data.
% Existing composition theorems show, how the privacy parameters $\epsilon$ and $\delta$ compose, when differentially private subroutines are combined.
The most 
%pessimistic
na\"ive
way is the \textit{linear} composition (Theorem 3.14 in \citet{Dwork14}), where the resulting parameter, which is often called \textit{cumulative privacy loss} (cumulative $\epsilon$ and $\delta$), are linearly summed up, $\epsilon = \sum_{t=1}^T\epsilon_t$ and $\delta = \sum_{t=1}^T\delta_t$ after the repeated use of data $T$ times with the per-use privacy loss $(\epsilon_t, \delta_t)$. 
Recently, \citet{2016arXiv160700133A} proposed the \textit{moments accountant} method, which provides an efficient way of combining $\epsilon$ and $\delta$ such that the resulting total privacy loss is significantly smaller than that by other composition methods (see \cref{supp:MA} for details).
%
% The moments accountant method takes advantage of the fact that when adding Gaussian noise in each training step, 
% the privacy loss in \cref{eq:privacyloss} also follows a Gaussian distribution. Hence, by observing the tail behaviour of the Gaussian random variable, one can obtain a tight moments bound which 
The resulting composition provides a better utility, meaning that a smaller amount of noise is required to add for the same privacy guarantee compared to other composition methods. 
% See \suppsecref{MA} for details. 
% Details are given in the appendix.

\paragraph{Random projections in the context of differential privacy}

A variant of our method involves projecting each input onto a lower-dimensional space using a \textit{Johnson-Lindenstrauss transform} (a.k.a., \textit{random projection}) \cite{privacy-via-the-johnson-lindenstrauss-transform}. We construct the projection matrix $\mR$ by drawing each entry from $\Nrm(0,1/D')$ where $D'$ is the dimension of the projected space.
This projection nearly preserves the distances between two points in the data space and in the embedding space, as this projection guarantees low-distortion embeddings. 
Random projections have previously been used to ensure DP \cite{Blocki2012TheJT}. However, here we only utilize them as a convenient method to reduce input dimension to our learnable linear maps. Since the random filters are data-independent, they do not need to be privatized.

% \section{Background on Interpretability}
% %
% Two key concepts regarding model interpretability are local and global explanations.
% A \emph{local explanation} of a model illustrates how the model behaves at and around a specific input, showing how relevant different features of the input were to the model decision.
% \emph{Global explanations} on the other hand aim to provide insight into how the model functions as a whole and, in the case of classification, what types of input the different classes are sensitive to.

% For many simple models like e.g. logistic regression both types of explanations are inherently available, as the shallow features for each class provide a simple interpretation of what the class is selecting for and each individual decision is based on how closely an input fits these features.  

% More complex models such as Deep Neural Networks tend to lack such inherent interpretability due to their nested feature structure. Attribution methods can provide local explanations by computing a linear approximation of the model at a given point in the input space. Such approximations can be seen as sensitivity maps that highlight, which parts of the input affect the model decision locally.

\section{Our method: Locally Linear Maps (LLM)}\label{sec:Meat_of_this_paper}

\paragraph{Motivation.}
As mentioned earlier, complex models such as deep neural networks tend to lack interpretability due to their nested feature structure. Gradient-based attribution methods can provide local explanations by computing a linear approximation of the model at a given point in the input space (see \secref{relatedwork} for more details). Such approximations can be seen as sensitivity maps that highlight which parts of the input affect the model decision locally. However, these approaches lack \emph{global} explanations that provide insight on how the model works as a whole, e.g., it is not straightforward to obtain class-wise key features. Furthermore, existing methods in the DP literature do not take into account the interpretability of learned models. In order to satisfy both interpretability and privacy demands, we desire a model with the following properties:
\begin{enumerate}
    \item It can provide both \textit{local and global explanations}.
    \item It has efficient ways to limit in the number of parameters to achieve a \textit{good privacy accuracy trade-off}.
    \item It is \textit{more expressive than standard linear models} to capture complex patterns in the data.
\end{enumerate}

\paragraph{Locally Linear Maps (LLM).} We introduce a set of local functions $f_k$ for each class $k$, and parameterize each $f_k$ by a combination of $M$ linear maps denoted by $g^k_m$. The $M$ linear maps are weighted separately for each class using the weighting coefficients $\sigma^k_m$, which determine how \textit{important} each linear map is for classifying a given input:
{\small\begin{align}
    f_k(\vx) &= \sum_{m=1}^M \sigma^k_m \; g^k_m (\vx),\\ 
    \mbox{where } g^k_m(\vx) &=\vw^k_m\trp\vx + \vb^k_m,
\end{align}}
{\small\begin{align}
    \text{and }  \sigma^k_m(\vx) &= \frac{\exp{\left[\beta \cdot g^k_m(\vx)  \right]}}{\sum_{m=1}^M \exp{\left[\beta \cdot g^k_m(\vx) \right]}.} \label{eq:weighting_coefficients}
\end{align}}
One way to choose the weighting coefficients is by assigning a probability to each linear map using the softmax function as in \cref{eq:weighting_coefficients}. We introduce a global inverse temperature parameter $\beta$ in the softmax to tune the sensitivity of the relative weighting -- large $\beta$ (small temperature) favors single filters; small $\beta$ (high temperature) favors several filters. The softmax weighting is useful for avoiding non-identifiability issues of parameters in mixture models. More importantly, the softmax weighting assigns an importance to each map particular to this example. In other words, it provides rankings of filters for different examples even if they are classified as the same class. We revisit this point in \secref{Experiments}.
We train the LLM by optimizing the following (standard) cross-entropy loss:
{\small\begin{align}\label{eq:NLL}
    \LL(\mW, \Dat) &= -\sum_{n=1}^N \sum_{k=1}^K y_{n,k} \log \hat{y}_{n,k}(\mW),
\end{align}} where we denote the parameters of LLM collectively by $\mW$, and we define the predictive class label by the mapping from the pre-activation through another softmax function.
{\small\begin{align}
\hat{y}_{n,k}(\mW) = \exp(f_k(\vx_n))/[\sum_{k'=1}^K \exp(f_{k'}(\vx_n))]
\end{align}}
When the number of filters per class is one, this reduces to logistic regression; increasing the number of filters adds expressive capacity to each class. The classification is approximately linear at the location of the input, which means that locally each model decision from a certain input can be explained using only the active filters, as we illustrate in the remainder of the paper. In addition the shallow nature of the model lends itself to global interpretability, as the filter-bank for each class is easily accessible and provides an overview of the inputs this class is sensitive to.

% In order to efficiently make this model private, we incorporate random projections to reduce the number of weights.

\paragraph{LLM as neural network approximations.}

One interpretation of LLM is as linearizations of neural networks.
Suppose we trained a neural network model on a $K$-class classification problem, where the network maps a high dimensional input $\vx \in \mathbb{R}^{D}$ to a class score function $\vs(\vx)$, i.e., the pre-activation before the final softmax, where $\vs(\vx)$ is a $K$-dimensional vector with entries $s_k$.
Denote the mapping $\vphi:\vx \mapsto \vs(\vx)$ and the parameters of the network by $\vtheta$. We would like to find the best approximation to the function $\vphi$, which presents interpretable features for classification and also guarantees a certain level of privacy.
For this, we take inspiration from gradient-based attribution methods for deep neural networks \cite{abs-1711-06104}.
These methods assume a set of attributions, at which the gradients of a classifier with respect to the input are maximized, \textit{and} that the gradient information provides interpretability as to why the classifier makes a certain prediction. 
More specifically, they consider a first order Taylor approximation of $\vphi$,
{\small\begin{align}
    \vphi(\vx) &\approx \vphi(\vx_0) + \vphi'(\vx_0)\trp(\vx-\vx_0) = \vphi'(\vx_0)\trp\vx + \bm{\alpha}, \nonumber 
\end{align}}
where $\vphi'(\vx_0)=\left[\frac{\partial}{\partial \vx}\vphi(\vx)\right]_{\vx=\vx_0}$, and shift term $\bm{\alpha}=\vphi(\vx_0) - \vphi'(\vx_0)\trp\vx_0$. 
% Notice that (a) the first order approximation is only accurate locally at $\vx_0$; 
% (b) a good location $\vx_0$ maximizes the gradient, because that point is intuitively where a tiny change in the input space would make the large change in the classification. Therefore, finding a good location $\vx_0$ and its gradient would reveal the most discriminative features for a given classifier. 
%
% (b) the gradient information at $\vx_0$  is useful as it reveals how a change in the input space would make a corresponding change in the classification outcome. 
Therefore, finding \textit{good} input locations $\vx_0$ to make the first order approximation to the function $\vphi$ and using their gradient information would reveal the discriminative features of a given classifier. 

There are two problems in directly using this approach. First, it is challenging to identify which input points (and how many of them) are  \textit{informative} to make an interpretable linear approximation of the classifier. 
Second, directly using $\vphi$ and its gradients violates privacy, as $\vphi$ contains sensitive information about individuals from the training dataset.
%
% As mentioned in \secref{background}, privatizing $\vphi$ requires computing the sensitivity, which determines an appropriate amount of noise to add. 
Privatizing $\vphi$ requires computing the sensitivity, which determines an appropriate amount of noise to add (see \secref{background}). 
In case of deep neural network models, we cannot analytically identify one datapoint's contribution to the learned function $\vphi$ appeared in \cref{eq:sensitivity}.
%
% to add often proves difficult because of its unknown sensitivity,
% such that we do not know how much noise to add to privatize $\vphi$. 
Thus, we cannot use the raw function $\vphi$ and its gradients, unless we privatize the parameters of $\vphi$\footnote{Once $\vphi$ is privatized, we can safely use post-processing methods for interpretability. A comparison to this is shown in \secref{Experiments}.}. 
% This typically incurs a high privacy loss to meet a certain level of classification accuracy as $\vtheta$ is typically high-dimensional. 
%
For these reasons, extracting a private approximation of $\vphi$ is difficult and we instead opt to train a model of the same form from scratch, leading us to LLMs, as described above. 

\begin{figure*}[htb]
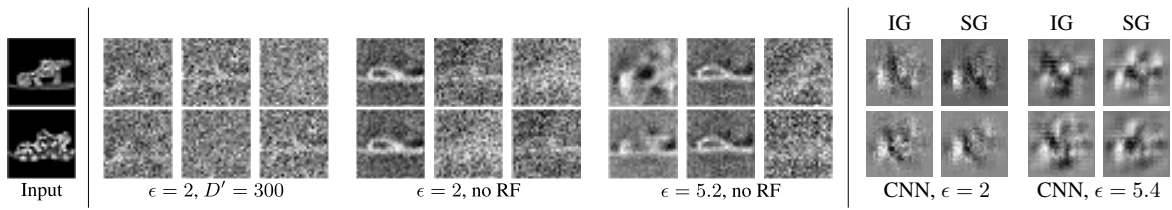

    \centering
    % \vspace{-0.4cm}
    \resizebox{0.9\textwidth}{!}{
    \includestandalone[mode=buildnew]{figures/rebuttal_plots/noise_privacy_filters}
    }
    \caption{
    Highest activated filters for 2 test inputs (left) under 3 DP setups, the leftmost filter having highest activation: the default setting with random filters at $D'=300$ and $\epsilon=2$ (center left),
    the same setting without random filters (center), and at lower privacy $\epsilon=5.2$ (center right). 
    Attribution plots from DP CNNs at matching privacy levels on the same input (right).
    % Removing random filters improves interpretability at the cost of utility. As privacy levels are reduced, filters and attributions become more interpretable. 
    }
    \label{fig:noisy_fmnist_filters}
\end{figure*}

\subsection{Differentially private LLM}

% \mpnote{@Frederik: Let's write DP stuff here. }
To produce differentially private LLM parameters $\widetilde\mW$, we adopt the moments accountant method combined with the gradient-perturbation technique \cite{2016arXiv160700133A}. This involves (i) perturbing gradients at each learning step when optimizing \cref{eq:NLL} for all LLM parameters $\mW$; and (ii) using the moments accountant method to compute the cumulative privacy loss after the training is over.  

When we perturb the gradient, we must ensure to add the right amount of noise. As there is no way of knowing how much change a single datapoint would make in the gradient's L2-norm, we rescale all the datapoint-wise gradients, $\vh_t(\vx_n) := \nabla_\mW \LL(\mW, \Dat_n)$ for all $n=\{1,\cdots,N\}$, by a pre-defined norm clipping threshold, $C$, as used in  \cite{2016arXiv160700133A}, i.e., 
% \begin{align}\label{eq:norm_clipping}
$\bar{\vh}_t(\vx_n) \leftarrow \vh_t(\vx_n)/\mbox{max}(1, \|\vh_t(\vx_i)\|_2/C)$.
% \end{align} 
\cref{algo:DP_LLM_Class} summarizes this procedure. We formally state that the resulting LLM are DP in theorem \ref{thm:dp_llm_class}.
\begin{algorithm}[h]
\caption{DP-LLM for interpretable classification}\label{algo:DP_LLM_Class}
\begin{algorithmic}
\vspace{0.1cm}
\REQUIRE Dataset $\Dat$, norm-clipping threshold $C$, privacy parameter $\sigma^2$, and learning rate $\eta_t$
\vspace{0.1cm}
\ENSURE $(\epsilon, \delta)$-DP locally linear maps for all $K$ classes, $\widetilde\mW$\\
\FOR{number of training steps $t \leq T$}
\STATE \textbf{1}: For each minibatch of size $L$, we noise up the gradient after clipping the norm of the datapoint-wise gradient 
%given in \cref{eq:norm_clipping} 
via 
${\tilde{\vh}_t \leftarrow \frac{1}{L} \left[ \sum_{n=1}^L\bar{\vh}_t(\vx_n) + \Nrm(0, \sigma^2 C^2 \mathbf{I}) \right].}$
\STATE \textbf{2}:  Then, we make a step in the descending direction by 
$\widetilde{\mW}_{t+1} \leftarrow \widetilde{\mW}_t - \eta_t \widetilde{\vh}_t$. 
 \ENDFOR
\STATE Calculate the cumulative privacy loss $(\epsilon, \delta)$ using the moments accountant.
\end{algorithmic}
\end{algorithm}
% \vskip-.5em

\begin{thm}\label{thm:dp_llm_class}
\algoref{DP_LLM_Class} produces  $(\epsilon, \delta)$-DP locally linear maps for all $K$ classes. 
\end{thm}
The proof is provided in \cref{supp:proof_thm1}. 

\paragraph{Improving privacy and accuracy trade-off under LLM}
For high-dimensional inputs such as images, we found that adding noise to the gradient corresponding to the full dimension of $\mW$ led to very low accuracies for private training. Therefore, we propose to incorporate the random projection matrix $\mR_m \in \mathbb{R}^{D' \times D}$ with  $D' \ll D$, which is shared among all classes $k$, to first decrease the dimensionality of the parameters that need to be privatized. Each LLM is therefore parameterized as 
%We now have the following parameterization for each locally linear maps, 
$\vw^k_m = \vm^k_m \mR_m$, where the effective parameter for each local linear map is $\vm^k_m \in \mathbb{R}^{D'}$. We perturb the gradient of $\vm^k_m$ for all $k$ and $m$ in each training step in \algoref{DP_LLM_Class} to produce DP linear maps, $\widetilde{\vw}^k_m = \widetilde\vm^k_m\mR_m$.

Due to the post-processing invariance property, we can use the differentially private LLM to make predictions on test data. 
Here we focus on guarding the training data's privacy and assume that the test data do not need to be privatized, which is a common assumption in DP literature.

\section{Experiments}\label{sec:Experiments}

In this section we evaluate the trade-off between accuracy, privacy, and interpretability for our LLM model on several datasets and compare to other methods where appropriate. Our implementation is available on GitHub\footnote{
%\url{github.com/frhrdr/dp-llm}
Link removed for anonymity
}.

\subsection{MNIST Classification}
\paragraph{Problem.} We consider the classification of MNIST \cite{lecun2010mnist} and Fashion-MNIST \cite{FashionMNIST} images with the usual train/test splits and train a  CNN\footnote{\url{github.com/pytorch/examples/blob/master/mnist/main.py}}
as a baseline model, which has two convolutional layers with 5x5 filters and first 20, then 50 channels each followed by max-pooling and finally a fully connected layer with 500 units. The model achieves 99\% test accuracy on MNIST and 87\% on Fashion-MNIST.

\paragraph{Setup.} We train several LLMs in the private and non-private setting. By default, we use LLM models with $M=30$ filters per class and random projections to $D'=300$ dimensions, which are optimized for 20 epochs using the Adam optimizer with learning rate 0.001, decreasing by 20\% every 5 epochs. On MNIST the model benefits from a decreased inverse softmax temperature $\beta=1/30$, while $\beta=1$ is optimal for Fashion-MNIST. We choose a large batch size of 500, as this improves the signal-to-noise ratio of our algorithm. In the private setting we clip the per-sample gradient norm to $C=0.001$ and train with $\sigma = 1.3$, which gives this model an $(\epsilon=2,\delta=10^{-5})$-DP guarantee via the moments accountant. For the low privacy regime $\epsilon \geq 4$ we train with a batch size of 1500 and for 60 epochs.

\begin{figure}[htb]
    \centering
    \includestandalone[mode=buildnew, scale=0.9]{figures/top_filters_fashion/top3_filters_fashion_2examples}
    \caption[]{Top 3 filters with associated weightings for test images from two classes. In most cases, a single filter dominates the softmax selection for the class.}
    \label{fig:mnist:top_filters_fashion_with_weights}
\end{figure}

\begin{figure}[htb]
    \centering
    \begin{tikzpicture}
    \node (Ac)
    {\includegraphics[trim=0 0 40 0, clip, scale=0.8]
    {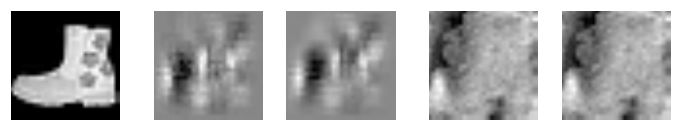}};
    \node[below=-0.2 of Ac] (Ba) {\includegraphics[trim=0 0 40 0, clip, scale=0.8]{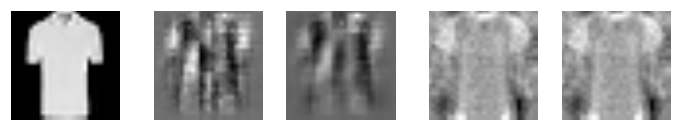}};
    \node[below=-0.2 of Ba] (Bb) {\includegraphics[trim=0 0 40 0, clip, scale=0.8]{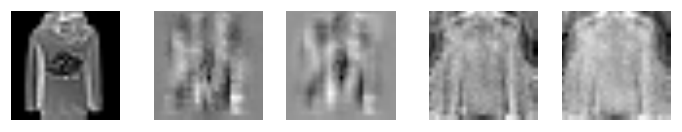}};
    \node[below=-0.2 of Bb] (Bc) {\includegraphics[trim=0 0 40 0, clip, scale=0.8]{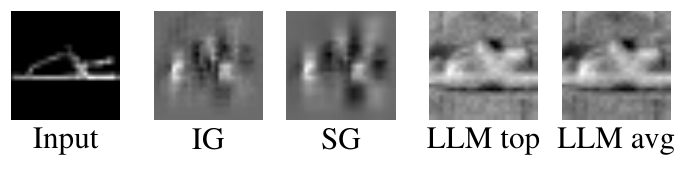}};
    \end{tikzpicture}
    \caption[]{Comparison of interpretability CNN and LLM. \textit{left to right}: input image, integrated gradient (IG) of CNN, smoothed gradient (SG) of CNN, and the top filter of LLM. The CNN attribution match well but aren't as easy to interpret as the simple filter.}
    \label{fig:mnist:ig_comparison}
\end{figure}

\begin{figure*}[htb]
    \centering
    % \includestandalone[mode=buildnew, scale=0.85]{figures/accuracy_vs_privacy/mnist}
    % \vskip-.75em
    \includegraphics[scale=0.85]{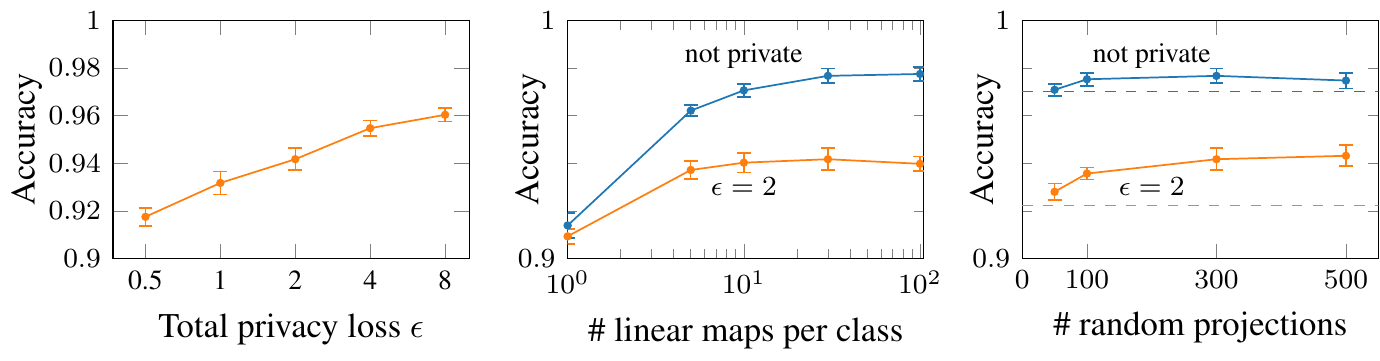}
    \caption[]{Accuracy of our LLM model on the MNIST testset for different levels of privacy and different 
    model configurations in the private (\protect\tikz[baseline=(current bounding box.base),inner sep=0pt]{\draw[line width=1.5pt, solid, C2] (0pt,3pt) --(15pt,3pt);}) and non-private
    (\protect\tikz[baseline=(current bounding box.base),inner sep=0pt]{\draw[line width=1.5pt, solid, C1] (0pt,3pt) --(15pt,3pt);}) setting. Errorbars are 2 stdev from $10$ random restarts; 
    dashed lines on the right (\protect\tikz[baseline=(current bounding box.base),inner sep=0pt] {\draw[line width=0.75pt, dashed, C1] (0pt,2pt) --(15pt,2pt); \draw[line width=0.75pt, dashed, C2] (0pt,5pt) --(15pt,5pt);}) 
    denote no random projections.} 
    \label{fig:mnist:accuracy_vs_privacy}
\end{figure*}

\paragraph{Inherent interpretability.} In order to highlight the interpretability of the LLM architecture, we compare learned filters of our model to two attribution methods applied to a neural network trained on the same data.
We train a simple CNN and an LLM on Fashion-MNIST to matching $87\%$ test accuracy and then visualize the CNN's sensitivity to test images using SmoothGrad \cite{SmilkovTKVW17} and integrated gradients \cite{SundararajanTY17} and compare these methods to LLM filters in \cref{fig:mnist:ig_comparison} (and \cref{app:fig:mnist:ig_comparison}).
Note that we do not multiply the integrated gradient with the input image, as Fashion-MNIST images have a mask-like effect which occludes the partial output of the method. We observe that both alternative attribution methods produce similar outputs, which are nonetheless hard to interpret, whereas the LLM filters show simplistic prototype images of the corresponding classes.
This is further illustrated in \cref{fig:mnist:top_filters_fashion_with_weights} where we show the three highest weighted filters for test images from three classes. The diversity of filters varies for different class labels, as some are more varied and harder to discriminate than others. For instance, while the sandal class (right) has filters which distinguish between different types of heels, 
%the ankle boot filters (right) show similar coarse features, which are sufficient for classifying a majority of the inputs correctly. The 
the coat filters (left) are mostly selective in the shoulder region and general silhouette, which is sufficient for classifying a majority of the inputs correctly, but some filters also track other features like arms, collar and zipper. 
The relevance weights for each filter show that in most cases, the top filter is assigned almost all the weight, indicating that the softmax is a good approximation of the maximum and the class features are indeed approximately linear locally.

\paragraph{Trade-offs with interpretability (\cref{fig:LoveTriangle} \circled{inner sep=1pt}{A} and \circled{inner sep=1pt}{B}).}
We investigate the learned LLM filters under increasing privacy guarantees and increased private utility as shown in \cref{fig:noisy_fmnist_filters}.
For two test inputs we plot the filters with highest activation in three DP setups. We compare the default setting with random filters at $D'=300$ and $\epsilon=2$, the same setting without random filters, and a lower privacy setting trained with $\epsilon=5.2$. 
The default setting optimizes privacy and utility at the expense of interpretability. As the figure shows, removing the random projections and reducing the level of privacy gradually restores interpretability of the filters. 
%When optimizing for accuracy in the high privacy case, however, we see that model interpretability suffers significantly. Removing the random projections causes a loss in test accuracy but yields clearer filters and equally, as the level of noise is reduced, more interpretable filters are retrieved. 
On the very right, we show attribution plots from CNNs trained with DP-SGD at $\epsilon=2$ and $\epsilon=5.4$ on the same input for reference. When comparing to \figref{mnist:ig_comparison}, one can see that the quality of the CNN attributions is diminished by the privacy constraints as well, to the point where it is hard to make out any connection to the input image.

\paragraph{Privacy vs. accuracy. (\cref{fig:LoveTriangle} \circled{inner sep=1pt}{C})}
In \cref{fig:mnist:accuracy_vs_privacy} (left) we show the trade-off of privacy strength and accuracy in our model. 
Note that current privatized network methods \cite{2016arXiv160700133A,PP_ConvDeepBeliefNet17} achieve an accuracy of 95\% for $\epsilon=2$ and up to 92\% for $\epsilon=0.5$, which is comparable to our mean accuracy of $94.2\pm 0.4\%$ and $91.8\pm 0.4\%$ respectively (on Fashion-MNIST we achieve $80.7\pm 0.6\%$ and $83.2\pm 0.4\%$).  
However, such a privatized network does not provide transparent explanations as opposed to our approach. Another popular reference model is the PATE method \cite{PATE2018}, which trains a student model to 98\% at $\epsilon=2$ on MNIST using an ensemble of teachers and additional public data. This is a special setting we don't consider here. The accuracy of the teacher votes alone lies at 94.4\% in the nonprivate setting, highlighting the importance of the additional data.
In the remainder of \cref{fig:mnist:accuracy_vs_privacy} we study the impact of varying the number of filters per class $M$ (center) and the output dimensionality of the random projections $D'$ (right) in private and non-private LLM models. Private LLMs deteriorate beyond a certain number of linear maps due to the increased noise needed to privatize them, whereas non-private models continue to benefit from additional filters. Increasing the dimensionality of the random projections benefits private training.

\subsection{Disease classification in a medical dataset}
%% so far just bullet points
\paragraph{Problem.} As a second task we consider disease classification in the Henan Renmin Hospital Data  \cite{li2017ensemble_medical_dataset_orig,Maxwell2017_medical_dataset}\footnote{downloaded from \url{http://pinfish.cs.usm.edu/dnn/}}. %\footnote{The dataset is provided by \protect\cite{Maxwell2017_medical_dataset} and was available at \url{http://pinfish.cs.usm.edu/dnn/}}.
It contains 110,300 medical records with 62 input features and 3 binary outputs. The input features are 4 basic examinations (sex, BMI, distolic, systolic), 26 items from blood examinations, 12 items from urine examinations, and 20 items from liver function tests. The three binary outputs denote %absence or presence of
three medical conditions -- hypertension, diabetes, and fatty liver -- which can also co-occur. 
Following \cite{Maxwell2017_medical_dataset} we transform this multi-label task into a multi-class problem by considering the powerset of the three binary choices as eight independent classes. Because these classes are highly imbalanced, we only retain the four most common classes, leaving us with 100,140 records.

\paragraph{Setup.} By default, we use an LLM model with $M=2$ filters per class and no random projections, which is optimized for 20 epochs using the Adam optimizer with learning rate 0.01, decreasing by 20\% every 5 epochs. We choose a batch size of 256. In the private setting we clip the per-sample gradient norm to $0.001$ and train with $\sigma = 1.25$, which gives this model an $(\epsilon\approx 1.5,\delta=2\cdot 10^{-5})$-DP guarantee via the moments accountant.

\begin{figure}[htb!]
% \vskip-0.5em
    \centering
    % \includestandalone[mode=buildnew]{figures/accuracy_vs_privacy/med_vertical}
    % \includestandalone[mode=buildnew, scale=0.8]{figures/accuracy_vs_privacy/med_vertical}
    % \vskip-0.5em
    \includegraphics[scale=0.8]{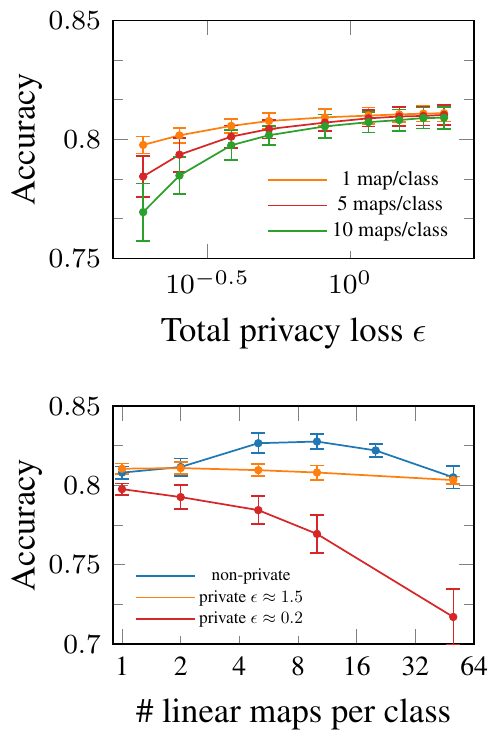}
    \caption[]{Accuracy of our LLM model on the Henan Renmin Hospital testset for different levels of privacy and different model configurations in the private and non-private setting. Errorbars are 2 stdev for $10$ random restarts.} 
    \label{fig:medical:accuracy_vs_privacy}
\end{figure}

We train a baseline DNN (3 dense hidden layers with 128 units each) as well as several LLMs with varying number of linear filters per class in private and non-private settings. In \cref{fig:medical:accuracy_vs_privacy} we visualize the trade-off between accuracy and privacy for varying privacy losses as well as numbers of linear maps. Like before, the accuracy deteriorates as we decrease the privacy loss (\cref{fig:medical:accuracy_vs_privacy} \textit{top}). As the number of linear maps per class is increased (\cref{fig:medical:accuracy_vs_privacy} \textit{bottom}), the accuracy for the private models also drops due to the privacy budget being spread across more parameters. We attribute the drop in performance for the non-private LLM with number of maps to optimization difficulties and local minima as well as higher sensitivity to hyperparameters. A small number of maps (between 2 and 5) is sufficient for this datasets, especially in the private setting. Our LLMs attain $82.8\pm 0.5\%$ (non-private), $82.0\pm 0.4\%$ ($\epsilon\approx 1.5$), and $79.8\pm 0.4\%$ ($\epsilon\approx 0.2$) compared to $84\pm 0.5\%$ for a non-private DNN.

\begin{figure}[htb]
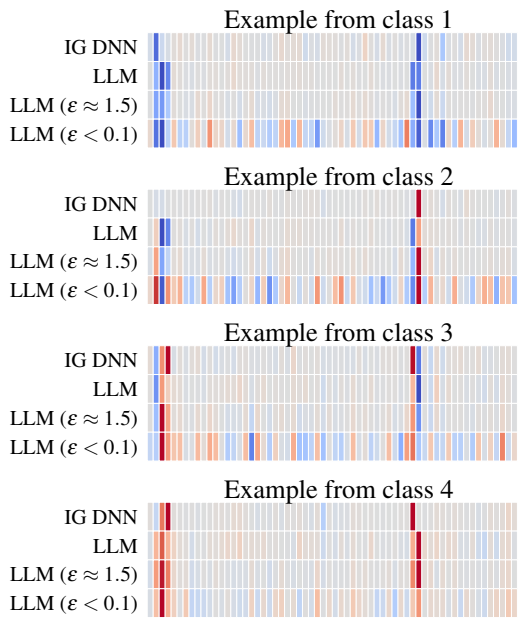

    \centering
    \includestandalone[mode=buildnew]{figures/med_interpret/med_plot_vertical}
    \caption[]{Integrated gradient (IG) and weighted linear filters (LLM; our method) for all 62 feature for one example from each class from the Henan Renmin dataset. For LMM we consider the non-private case (LLM) as well as two private cases with strong ($\epsilon<0.1$) and weaker ($\epsilon\approx 1.5$) privacy. Entries are normalized and colorcoded between \protect\tikz[baseline=(current bounding box.base),inner sep=0pt]{\draw[line width=3pt, solid, darkblue] (0pt,3pt) --(10pt,3pt);}${}=-1$, \protect\tikz[baseline=(current bounding box.base),inner sep=0pt]{\draw[line width=3pt, solid, lightgray] (0pt,3pt) --(10pt,3pt);}${}=0$, and \protect\tikz[baseline=(current bounding box.base),inner sep=0pt]{\draw[line width=3pt, solid, darkred] (0pt,3pt) --(10pt,3pt);}${}=1$.}
    \label{fig:medical:interpretability}
\end{figure}

In \cref{fig:medical:interpretability} we consider an example from each class and show the weighted linear maps by the LLM for each example as well as its integrated gradients (IG) \cite{SundararajanTY17}. For our LLM we consider the non-private and two private cases. In general, there is good agreement between all attribution methods; they are relatively sparse and focus on a small set of features. We found that IG varied much more between examples from the same class than our LLM (see \cref{app:fig:medical:interpretability} in \cref{app:sec:experiments_attribution_medical}). For strong privacy ($\epsilon < 0.1$), the linear maps are much less sparse, highlighting the trade-off between interpretability and privacy.

\section{Related Work}\label{sec:relatedwork}

\paragraph{Interpretability.} The \textit{saliency} map and \textit{gradient-based attribution} methods are one of the most popular explanation methods that identify relevant regions and assign importance to each input feature (e.g., pixel for image data) \cite{SelvarajuDVCPB16,Ribeiro2016,SmilkovTKVW17,SundararajanTY17,MontavonBBSM15,journal.pone.0130140,abs-1711-06104}. 
% Beyond the input feature level, several recent works attempt to identify high-level concepts that humans easily understand from the learned models (e.g., \cite{BauZKOT17, 2017arXiv171111279K} among many). In this paper, we focus on the methods for interpretability of input-level features. 
%
These methods typically use first-order gradient information of a complex model with respect to inputs, to produce maps that indicate the relative importance of the different input features for the classification. 
An obvious downside of these approaches is that they provide explanations conditioned on only \emph{a single} input %, or one baseline input, 
and hence it is necessary to manually assess each input of interest in order to draw a class-wide conclusion. 
In contrast, our approach can draw class-wide conclusions without manually assessing each input, because it outputs the most relevant explanations in terms of a collection of linear maps for each class. 
For explanations conditioned on any specific input, our model can provide an input-dependent weighted collection of these features related to that specific input. 

\paragraph{Privacy.} To privatize complex models, such as deep neural networks, a popular approach is to add noise to the gradients in the stochastic gradient descent (SGD) algorithm \cite{2016arXiv160700133A,papernot:private-training,DBLP:journals/corr/abs-1710-06963}.
An alternative approach is to directly perturb the objective with additive noise
\cite{Obj_Pertur_Taylor_12,DP_DeepAutoEnc,PP_ConvDeepBeliefNet17}. In these works, the objective function is approximated by the Taylor expansion, and the resulting coefficients of the polynomials are perturbed before training. 
We found the latter approach less practical than the former, as 
we need to choose which order of polynomial degree to use. 
%Typically, adding more layers introduces a more nested-ness in the objective function in which case using a higher order approximation is more suitable to approximate the loss function accurately. 
Typically, adding more layers introduces a more nested-ness in the objective function requiring a higher order polynomial for accurate approximation. 
A high degree of polynomial approximation, however, increases the privacy loss as the dimensionality of the coefficients grow. 
%
% From our perspective, the gradient perturbation method is simple to use and model agnostic, and there are many successful examples of the gradient perturbation methods used in slightly different settings than privatizing a whole model from scratch, such as knowledge transfer from teacher models to student models in \cite{papernot:private-training,PATE2018,2016arXiv160700133A}. 
% NOTE (frederik): this is not correct, PATE is not a gradient perturbation method
%
From our perspective, the gradient perturbation method is simple to use and model agnostic.
Recently proposed methods for private training through ensembles of teacher models \cite{papernot:private-training,PATE2018} are less useful to us here, as they consider a special setting where some non-private data is available in addition to the private dataset. 

Our method distinguishes itself by making interpretability a key component of the trained model and does not rely on access to additional public data. 
%However, none of these methods took interpretability into account, and some of the work assume the availability of public data to train a significant part of their model to decrease the necessary privacy budget to train the entire model. In our method, no access to public data is assumed and interpretability through linear maps is a key component of the trained model. 

\paragraph{Mixtures of Experts.} Our LLMs are reminiscent of \emph{Mixture of experts} (ME) models. MEs assign different specialized linear models to different parts of input space in a discriminative task %(\citet{Masoudnia2014_mixture_of_experts} provide an overview of existing ME models).
(see \citet{Masoudnia2014_mixture_of_experts} for an overview of existing ME models). 
In our case, each local expert model is class specific and contributes to a weighted linear map for that class. The weighting provides an input-dependent \emph{significance} for each linear map, and considering more than one map per class increases flexibility to fit the data better.
% Another relevant model is the \textit{Mixture of factor analyzer} (MFA), which has a very similar flavor as the ME models, but developed for density estimation of high-dimensional real-valued data \cite{Ghahramani97theem}. 
%\mbnote{what's the relevance of MFA? doesn't become clear. "Mixtures of factor analyzer (MFA) are very similar to ME models but have been developed for ..." }
\textit{Mixtures of factor analyzers} (MFA) are also similar to ME models but have been developed for density estimation of high-dimensional real-valued data \cite{Ghahramani97theem}

\section{Conclusion and Discussion}
We proposed a family of simple models which
uses several \emph{locally linear maps} (LLM) per class to provide interpretable features in a privacy-preserving manner while maintaining high classification accuracy. 
Results on two image benchmark datasets as well as a medical dataset indicate that a reasonable trade-off between classification accuracy, privacy \emph{and} interpretability can indeed be struck and tuned by varying the number of linear maps.
Nevertheless, several open questions for future research remain. 
First, the datasets in this paper are still relatively simple, such that it would be intriguing to see the limits of complexity the LLM model can model with a  sufficiently high accuracy.
Second, the current model does not interact with a larger and richer counterpart, such as a neural network, due to privacy constraints. 
It would be interesting to investigate if gaining gradient information of a more flexible model at particularly important input points in a DP way would be possible, in order to combine benefits of both models.
%Second, the current model does not interact with a larger and richer counterpart, such as a neural network, due to privacy constraints. It would be interesting to investigate if gaining gradient information of a more flexible model at particularly important input points in a differentially private way would be possible, in order to combine benefits of both models.
% Lastly, 

% \clearpage
% \small
% \bibliography{ref_priv_pred}
% \bibliographystyle{abbrv}

% \subsection*{Acknowledgements}
% The work of M. Park and F. Harder is supported by the Max Planck Society, as well as the 
% Gibs Sch{\"u}le Foundation and the Institutional Strategy of the University of T{\"u}bingen (ZUK63).
% %
% M. Bauer gratefully acknowledges partial funding through a Qualcomm studentship in technology as well as by the EPSRC. 
% %
% We would like to thank Maryam Ahmed for useful comments on the manuscript, Patrick Putzky and Wittawat Jitkritum for their scientific insights, and Been Kim and Isabel Valera for their inputs on interpretability research.  Lastly, M. Park is grateful to Gilles Barthe for the inspiration of studying the trade-off between interpretability and privacy. 

\subsection*{Acknowledgments}
M. Park and F. Harder are supported by the Max Planck Society and the Gibs Sch{\"u}le Foundation and the Institutional Strategy of the University of T{\"u}bingen (ZUK63).
F. Harder thanks the International Max Planck Research School for Intelligent Systems (IMPRS-IS) for its support.
M. Bauer gratefully acknowledges partial funding through a Qualcomm studentship in technology as well as by the EPSRC. 
We thank Maryam Ahmed for useful comments on the manuscript, Patrick Putzky and Wittawat Jitkritum for their scientific insights, and Been Kim and Isabel Valera for their inputs on interpretability research.  
M. Park is grateful to Gilles Barthe for the inspiration of studying the trade-off between interpretability and privacy. 

\FloatBarrier
\newpage
~
\newpage

\bibliography{ms.bib}
\bibliographystyle{aaai} 

\newpage
\appendix
\textbf{\LARGE \centering Supplementary Material}

\section{Properties of differential privacy}\label{supp:DPproperties}
\paragraph{Post-processing invariance}
The following proposition states the post-processing invariance property of differential privacy. 
\begin{prop}[Proposition 2.1 \citep{Dwork14}] \label{prop:postprocessing_invariance}
Let a mechanism that maps data where $\chi$ is the data universe to an output space, i.e., $\mathcal{M} : \mathbb{N}^{|\chi|} \mapsto \mathcal{R}$ 
be a randomized algorithm that is $(\epsilon, \delta)$-differentially private. Let $f : \mathcal{R} \mapsto \mathcal{R}'$ be an arbitrary, data-independent, randomized mapping. 
Then $f \circ \mathcal{M} : \mathbb{N}^{|\chi|} \mapsto \mathcal{R}'$ is also $(\epsilon, \delta)$- differentially private.
\end{prop}

\section{A short summary for moments accountant method}\label{supp:MA}

\paragraph{The moments accountant.}
The privacy loss in \eqref{privacyloss} is a random variable once we add noise to the output of the algorithm. In fact, when we add Gaussian noise, the privacy loss random variable is also Gaussian distributed. Using the tail bound of Gaussian privacy loss random variable, the \textit{moments accountant} method \cite{2016arXiv160700133A} provides a clever way of combining $\epsilon$ and $\delta$ such that the resulting total privacy loss is significantly smaller than other composition methods.

% Here we provide a short summary of the method. 

In the \textit{moments accountant} method, the cumulative privacy loss is calculated by bounding the moments of the privacy loss random variable $L^{(o)}$. 
First off, each $\lambda$-th moment, where $\lambda$ can be positive integers, is defined as the log of the moment generating function evaluated at $\lambda$, i.e., 
%\begin{equation}
%\label{eq:Moment}
$\alpha_{\mathcal{M}}(\lambda; \Dat, \Dat') = \log \mathbb{E}_{o \sim \mathcal{M}(\Dat)} \left[ e^{\lambda  L^{(o)}}\right].$
%\end{equation}
%Note that this is different from the usual calculation of $\lambda$-th moment: $m_\lambda = \mathbb{E}[o^\lambda] = \frac{d^\lambda}{d t^\lambda} \mathbb{E}[\text{e}^{to}]$,  evaluated at ${t=0}$. 
Then, by taking the maximum over the neighboring datasets, we compute the worst case $\lambda$-th moment by, $ \alpha_{\mathcal{M}}(\lambda) = \max_{\Dat, \Dat'}\alpha_{\mathcal{M}}(\lambda; \Dat, \Dat')$, where the form of $\alpha_{\mathcal{M}}(\lambda)$ is determined by the moment of a Gaussian random variable. 
The moments accountant then computes $\alpha_{\mathcal{M}}(\lambda) $ at each step.
The composability theorem (Theorem 2.1 in \cite{2016arXiv160700133A}) states that
 the $\lambda$-th moment 
 %in \eqref{Moment}
 composes linearly if we add independent noise at each training step. So, we can simply sum up the upper bound on each $\alpha_{\mathcal{M}_t}$ to obtain an upper bound on the total $\lambda$-th moment after $T$ compositions, 
%\begin{equation}
%\label{eq:Composibility}
$\alpha_{\mathcal{M}}(\lambda) \leq  \sum_{t=1}^T \alpha_{\mathcal{M}_t}(\lambda).$ %(\lambda)'s added to this equation by Jimmy, as this is how it's written in Abadi et al (2016), and since \alpha_{\mathcal{M}} is otherwise ambiguously defined.
%\end{equation} 
%
Finally, once the moment bound is computed, we can convert the $\lambda$-th moment to the ($\epsilon,\delta$)-DP guarantee by, 
%\begin{equation}
%\label{eq:tail_bound}
$\delta = \min_{\lambda} \exp \left[ \alpha_{\mathcal{M}}(\lambda) - \lambda \epsilon \right]$, for any $\epsilon>0$. See Appendix A in \cite{2016arXiv160700133A} for the proof.
%\end{equation} 

\section{Proof of Theorem 1}\label{supp:proof_thm1}
\begin{proof}
We first prove that one gradient step in \algoref{DP_LLM_Class} produces differentially private locally linear maps, then generalize this result for the $T$ number of gradient steps. 

Given an initial \textit{data-independent} value of $\mW_0$, if we add Gaussian noise to the norm-clipped gradient evaluated on the subsampled data with the samping rate $q=L/N$, then due to the \textit{Gaussian mechanism} (Theorem 3.22 in \cite{Dwork14}) and Theorem 1 in \cite{2016arXiv160700133A}, 
the resulting estimate $\tilde{\mW}_1$ from a single gradient step (i.e., the step $\textbf{2}$ in \algoref{DP_LLM_Class}) is $(\epsilon', \delta')$-differentially private, where $\sigma \geq c \cdot q\sqrt{\log(1/\delta')}/\epsilon'$ with some constant $c$. 
Now, as $\tilde{\mW}_1$ is already privatized, we can make further gradient steps from $\tilde{\mW}_1$, which makes $\tilde\mW_2$ also  $(\epsilon', \delta')$-differentially private, as the only part that depends on the data is the gradient which we perturb for privacy. Applying the same amount of Gaussian noise to the gradient in each step ensures each $\tilde\mW_t$ for all $t$ also  $(\epsilon', \delta')$-differentially private. 

Finally, the composibility and tail bound in Theorem 2 in \cite{2016arXiv160700133A} proves that the cumulative privacy loss after $T$ training steps computed by the moments accountant method ensures ($\epsilon, \delta$)\footnote{Note that we can identify the exact relationship between the cumulative loss ($\epsilon, \delta$) and $(\epsilon', \delta')$ numerically only, due to the constant factor in $\sigma \geq c \cdot q\sqrt{\log(1/\delta')}/\epsilon'$. We use code published by \cite{2016arXiv160700133A} to compute these numerically. }-DP locally linear maps. 
\end{proof}

% \newpage
\section{Additional Experimental Results}

\subsection{Attribution methods on private and nonprivate CNNs}

\begin{figure}[htb]
    \centering
    \includestandalone[mode=buildnew, scale=1.2]{figures/rebuttal_plots/cnn-attr}
    \caption[]{Gradient attributions for two fashion Mnist samples shown on the left using two different attribution methods: Integrated Gradient and Smoothed Gradient. We test two CNNs of different size with 2 convolutional and 2 dense layers. $M_a$ has 431080 parameters,
    % \footnote{$M_a$ source code: 
    % \url{ https://github.com/pytorch/examples/blob/master/mnist/main.py}
    % }
    while $M_b$ is designed for privacy and has only 26010 parameters.
    % \footnote{$M_b$ source code: 
    % \url{https://github.com/tensorflow/privacy/blob/master/tutorials/mnist_dpsgd_tutorial.py}
    % }
    Each model is trained privately to $\varepsilon = 1.19$ and non-privately (non-p). We observe that both model choice and privacy constraints impact the quality of the attributions. While the DP version of $M_b$ achieves 81\% classification accuracy, $M_a$ only reaches 75\% in the private setting, so the interpretability/privacy/accuracy trade-off is already relevant at the point of architecture selection.
    }
    \label{fig:mnist:cnn_attributions}
\end{figure}

\newpage

\subsection{Comparison of attribution methods on MNIST}

\begin{figure}[htb]
    \centering
\includegraphics[scale=0.9]{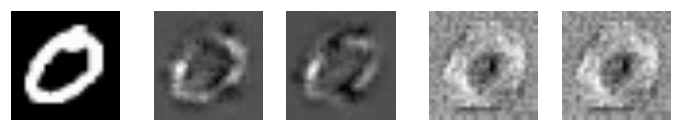}\\
\includegraphics[scale=0.9]{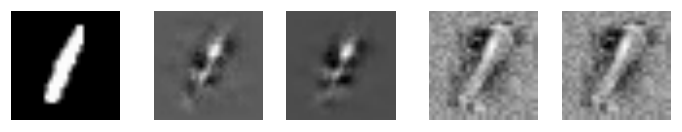}\\
 \includegraphics[scale=0.9]{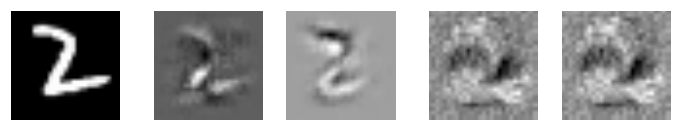}\\
\includegraphics[scale=0.9]{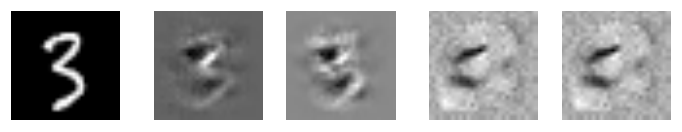}\\
\includegraphics[scale=0.9]{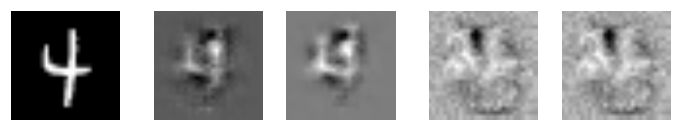}\\
\includegraphics[scale=0.9]{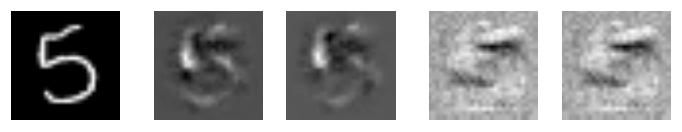}\\
\includegraphics[scale=0.9]{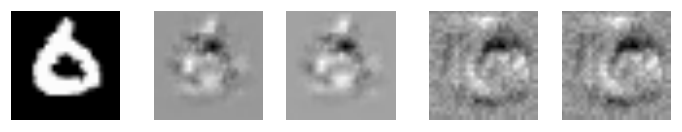}\\
\includegraphics[scale=0.9]{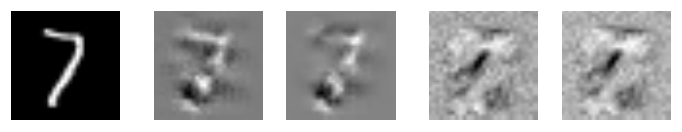}\\
\includegraphics[scale=0.9]{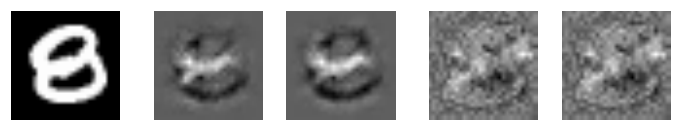}\\
\includegraphics[scale=0.9]{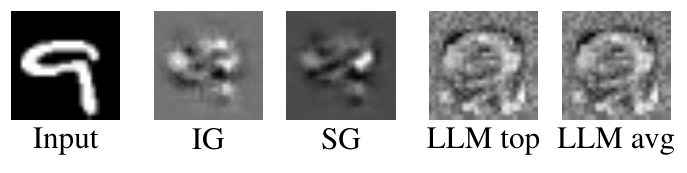}
\caption{
Comparison between different attribution methods, similar to \cref{fig:mnist:ig_comparison} but for MNIST, showing integrated gradient (IG) and smoothed gradient (SG) for a CNN and the top filter and weighted filter average for the selected class of an LLM. As most weight is assigned to the top filter, the weighted average looks almost identical. On this dataset, the network attributions resemble the input more closely than on Fashion-MNIST, highlighting relevant edges. LLM filters exhibit the same kind of coarse prototypical images with pronounced edges as on Fashion-MNIST.
}
\label{app:fig:mnist:ig_comparison}
\end{figure}

\newpage

\subsection{Attribution methods on medical data}
\label{app:sec:experiments_attribution_medical}

\begin{figure}[h!]
    \centering
    \begin{tikzpicture}
    \node(left1) {\includegraphics{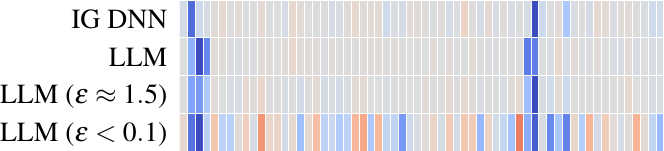}};
    \node[below=0.1 of left1] (left2) {\includegraphics{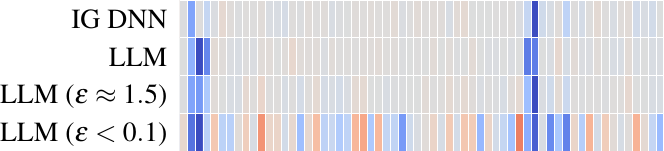}};
    \node[below=0.1 of left2] (left3) {\includegraphics{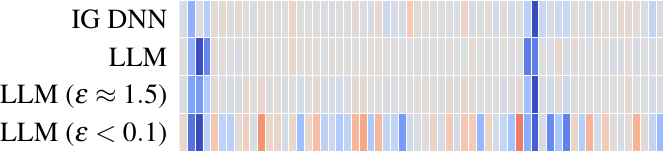}};
    \node[below=0.1 of left3] (left4) {\includegraphics{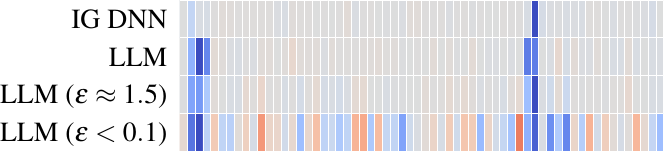}};
    % \node[right=0.5 of left1](right1) {\includegraphics{figures/med_interpret/compare_ig_weights_6_class1.pdf}};
    % \node[below=0.1 of right1] (right2) {\includegraphics{figures/med_interpret/compare_ig_weights_9_class1.pdf}};
    % \node[below=0.1 of right2] (right3) {\includegraphics{figures/med_interpret/compare_ig_weights_49_class1.pdf}};
    % \node[below=0.1 of right3] (right4) {\includegraphics{figures/med_interpret/compare_ig_weights_28_class1.pdf}};
    \node[anchor=center] at ($(left1.north)+(1,0.02)$) {\textbf{Examples from class 1}};
    % \node[anchor=center] at ($(right1.north)+(1,0.02)$) {\textbf{Example from class 2}};
    \end{tikzpicture}
    
    \vskip 0.5em
    
    \begin{tikzpicture}
    \node(left1) {\includegraphics{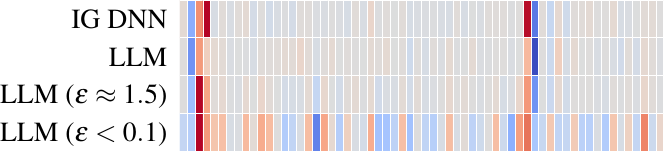}};
    \node[below=0.1 of left1] (left2) {\includegraphics{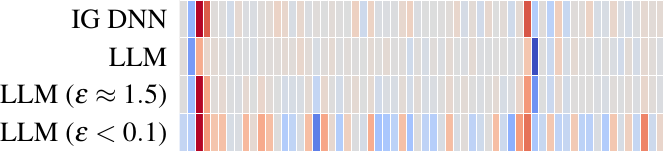}};
    \node[below=0.1 of left2] (left3) {\includegraphics{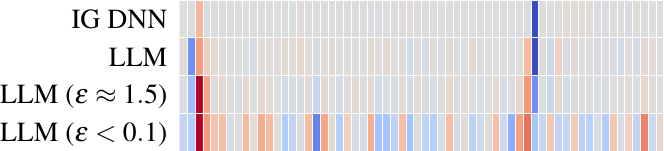}};
    \node[below=0.1 of left3] (left4) {\includegraphics{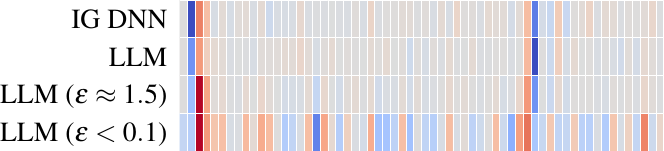}};
    
    %  \node[right=0.5 of left1](right1) {\includegraphics{figures/med_interpret/compare_ig_weights_5_class3.pdf}};
    % \node[below=0.1 of right1] (right2) {\includegraphics{figures/med_interpret/compare_ig_weights_15_class3.pdf}};
    % \node[below=0.1 of right2] (right3) {\includegraphics{figures/med_interpret/compare_ig_weights_16_class3.pdf}};
    % \node[below=0.1 of right3] (right4) {\includegraphics{figures/med_interpret/compare_ig_weights_26_class3.pdf}};
    \node[anchor=center] at ($(left1.north)+(1,0.02)$) {\textbf{Examples from class 3}};
    % \node[anchor=center] at ($(right1.north)+(1,0.02)$) {\textbf{Example from class 4}};
    % \node[anchor=center] at ($(leftbottom.north)+(1,0.02)$) {Example from class 3};
    % \node[anchor=center] at ($(rightbottom.north)+(1,0.02)$) {Example from class 4};
    \end{tikzpicture}
    
    \caption[]{Integrated gradient (IG) and weighted linear filters (LLM; our method) for all 62 feature for four example from 2 classes from the Henan Renmin dataset. For LMM we consider the non-private case (LLM) as well as two private cases with strong ($\epsilon<0.1$) and weaker ($\epsilon\approx 1.5$) privacy. Entries are normalized and colorcoded between \protect\tikz[baseline=(current bounding box.base),inner sep=0pt]{\draw[line width=3pt, solid, darkblue] (0pt,3pt) --(10pt,3pt);}${}=-1$, \protect\tikz[baseline=(current bounding box.base),inner sep=0pt]{\draw[line width=3pt, solid, lightgray] (0pt,3pt) --(10pt,3pt);}${}=0$, and \protect\tikz[baseline=(current bounding box.base),inner sep=0pt]{\draw[line width=3pt, solid, darkred] (0pt,3pt) --(10pt,3pt);}${}=1$. This is an extended version of \cref{fig:medical:interpretability}. Note that there is less variability between explanations/attributions for LLM (non-private) than there is for integrated gradients.}
    \label{app:fig:medical:interpretability}
\end{figure}

\end{document}

%% file: figures/rebuttal_plots/noise_privacy_filters.tex
    \begin{tikzpicture}
    \node (big) {\includegraphics{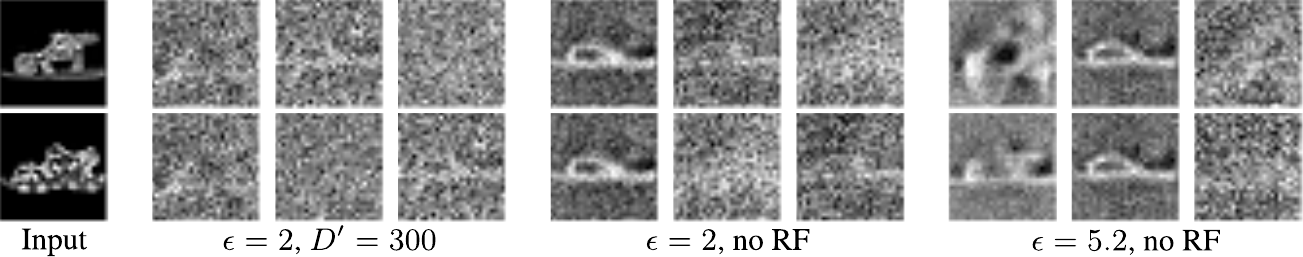}};
    
    \node[above right=0.75 and 0.3 of big.east, anchor=west] (IGs1eps2) {\includegraphics[width=1.05cm]{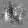}};
    \node[right= 0. of big.east] (tmp) {};
    \node[right=-0.1 of IGs1eps2] (SGs1eps2) {\includegraphics[width=1.05cm]{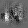}};
    \node[below right=0.4 and 0.3 of big.east, anchor=west] (IGs2eps2) {\includegraphics[width=1.05cm]{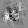}};
    \node[right=-0.1 of IGs2eps2] (SGs2eps2) {\includegraphics[width=1.05cm]{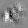}};
    \node[above=-0.1 of IGs1eps2, anchor=south] (IGtext) {IG};
    \node[above=-0.1 of SGs1eps2, anchor=south] {SG};
    \node[anchor=center] at ($(IGs2eps2.south)!0.5!(SGs2eps2.south) - (0,0.1)$) (label) {CNN, $\epsilon=2$};
    \draw (tmp|-IGtext.north) --(tmp|-label.south);
    
    \node[right=1.3 of big.west] (tmpc){};
    \draw (tmpc|-IGtext.north) --(tmpc|-label.south);
    
    \node[right=0.1 of SGs1eps2] (IGs1eps5) {\includegraphics[width=1.05cm]{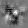}};
    \node[right=-0.1 of IGs1eps5] (SGs1eps5) {\includegraphics[width=1.05cm]{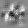}};
    \node[right=0.1 of SGs2eps2] (IGs2eps5) {\includegraphics[width=1.05cm]{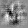}};
    \node[right=-0.1 of IGs2eps5] (SGs2eps5) {\includegraphics[width=1.05cm]{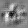}};
    \node[above=-0.1 of IGs1eps5, anchor=south] (IGtext) {IG};
    \node[above=-0.1 of SGs1eps5, anchor=south] {SG};
    \node[anchor=center] at ($(IGs2eps5.south)!0.5!(SGs2eps5.south) - (0,0.1)$) (label) {CNN, $\epsilon=5.4$};
    \end{tikzpicture}
    

%% file: figures/top_filters_fashion/top3_filters_fashion_2examples.tex
\begin{tikzpicture}
    \node (Aa) 
    {\includegraphics[scale=0.75]{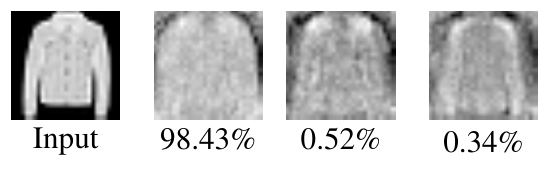}};
    \node[below=-0.3 of Aa] (Ab) {\includegraphics[scale=0.75]{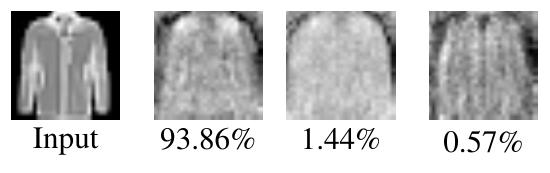}};
    \node[below=-0.3 of Ab] (Ac) {\includegraphics[scale=0.75]{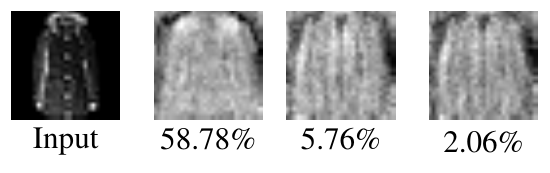}};
    \node[right=0.2 of Aa] (Ba) {\includegraphics[scale=0.75]{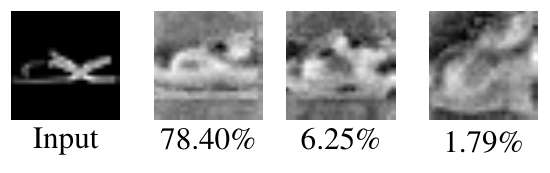}};
    \node[below=-0.3 of Ba] (Bb) {\includegraphics[scale=0.75]{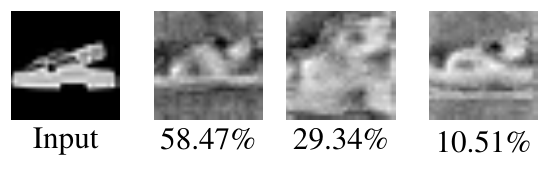}};
    \node[below=-0.3 of Bb] (Bc) {\includegraphics[scale=0.75]{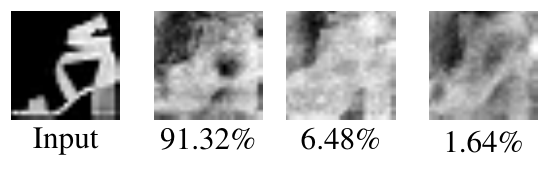}};
    % \node[above=0.05  of Aa, anchor=center] {MNIST};
    % \node[above=0.05  of Ba, anchor=center] {fashion MNIST};
\end{tikzpicture}

%% file: figures/med_interpret/med_plot_vertical.tex
\begin{tikzpicture}
\node(ex1) {\includegraphics{figures/med_interpret/compare_ig_weights_0_class0.pdf}};
\node[below = 0.3 of ex1] (ex2){\includegraphics{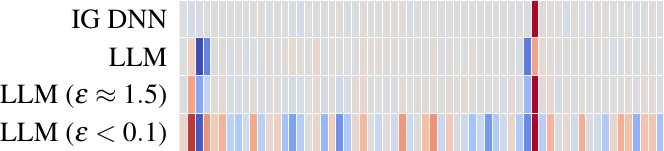}};
\node[below=0.3 of ex2] (ex3) {\includegraphics{figures/med_interpret/compare_ig_weights_55_class2.pdf}};
\node[below = 0.3 of ex3] (ex4)  {\includegraphics{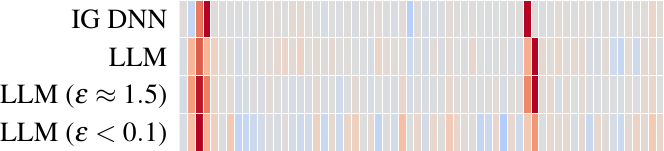}};
\node[anchor=center] at ($(ex1.north)+(1,0.02)$) {Example from class 1};
\node[anchor=center] at ($(ex2.north)+(1,0.02)$) {Example from class 2};
\node[anchor=center] at ($(ex3.north)+(1,0.02)$) {Example from class 3};
\node[anchor=center] at ($(ex4.north)+(1,0.02)$) {Example from class 4};
\end{tikzpicture}

%% file: figures/rebuttal_plots/cnn-attr.tex
\begin{tikzpicture}
    \node (s1bni)  %Sandal 1 Big Nonprivate Integrated
    {\includegraphics[scale=0.9]{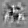}};
    \node[below=-0.2 of s1bni] (s1bpi)  %Sandal 1 Big Private Integrated
    {\includegraphics[scale=0.9]{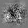}};
    \node[right=-0.2 of s1bni]  (s1sni)  %Sandal 1 Small Nonprivate Integrated
    {\includegraphics[scale=0.9]{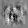}};
    \node[right=-0.2 of s1bpi] (s1spi)  %Sandal 1 Small Private Integrated
    {\includegraphics[scale=0.9]{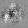}};

    \node[right=0.2 of s1sni] (s1bns)  %Sandal 1 Big Nonprivate Smooth 
    {\includegraphics[scale=0.9]{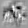}};
    \node[below=-0.2 of s1bns] (s1bps)  %Sandal 1 Big Private Smooth 
    {\includegraphics[scale=0.9]{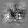}};
    \node[right=-0.2 of s1bns] (s1sns)  %Sandal 1 Small Nonprivate Smooth 
    {\includegraphics[scale=0.9]{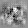}};
    \node[right=-0.2 of s1bps] (s1sps)  %Sandal 1 Small Private Smooth 
    {\includegraphics[scale=0.9]{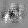}};

    \node[below=0.2 of s1bpi] (s2bni) {\includegraphics[scale=0.9]{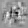}};
    \node[below=-0.2 of s2bni] (s2bpi) {\includegraphics[scale=0.9]{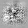}};
    \node[right=-0.2 of s2bni] (s2sni) {\includegraphics[scale=0.9]{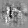}};
    \node[right=-0.2 of s2bpi] (s2spi) {\includegraphics[scale=0.9]{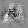}};

    \node[right=0.2 of s2sni] (s2bns) {\includegraphics[scale=0.9]{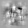}};
    \node[below=-0.2 of s2bns] (s2bps) {\includegraphics[scale=0.9]{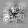}};
    \node[right=-0.2 of s2bns] (s2sns) {\includegraphics[scale=0.9]{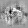}};
    \node[right=-0.2 of s2bps] (s2sps) {\includegraphics[scale=0.9]{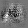}};
    
    \node (s1) at ($(s1bni.west) + (-0.35,-0.34)$) {\includegraphics[scale=0.9]{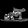}};
    
    \node (s2) at ($(s2bni.west) + (-0.35,-0.34)$) {\includegraphics[scale=0.9]{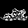}};
    
    \node[above=0 of s1sni] {\small $M_b$};
    \node[above=0 of s1bni] {\small $M_a$};
    \node[above=0 of s1sns] {\small $M_b$};
    \node[above=0 of s1bns] {\small $M_a$};
    
    \node[right=0 of s1sns] {\small non-p.};
    \node[right=0 of s1sps] {\small $\varepsilon = 1.19$};
    \node[right=0 of s2sns] {\small non-p.};
    \node[right=0 of s2sps] {\small $\varepsilon  = 1.19$};
    
    \node at ($(s2bpi.south) + (0.35,-0.2)$) {\small Integrated};
    \node at ($(s2bps.south) + (0.35,-0.2)$)  {\small Smoothed};
    
    %
    %\node[anchor=center] at ($(A.north) + (0,0.5)$) {\small no projections, vary $\epsilon$};
    %\node[above=0 of B, anchor=center] {\small non-private, vary $d'$};
    %\node[above=0 of C, anchor=center] {\small $d'=300$ projections, vary $\epsilon$};
    
\end{tikzpicture}